# Cybersickness Detection through Head Movement Patterns: A Promising Approach

Masoud Salehi[1*], Nikoo Javadpour[1*], Brietta Beisner[1], Mohammadamin Sanaei[1], Stephen B. Gilbert[1]

[1] Iowa State University, Ames IA 50011, USA

**Abstract.** Despite the widespread adoption of Virtual Reality (VR) technology, cybersickness remains a barrier for some users. This research investigates head movement patterns as a novel physiological marker for cybersickness detection. Unlike traditional markers, head movements provide a continuous, non-invasive measure that can be easily captured through the sensors embedded in all commercial VR headsets.

We used a publicly available dataset from a VR experiment involving 75 participants and analyzed head movements across six axes (up/down, left/right, forward/backward, and rotational movements). An extensive feature extraction process was then performed on the head movement dataset and its derivatives, including velocity, acceleration, and jerk. Three categories of features were extracted, encompassing statistical, temporal, and spectral features. Subsequently, we employed the Recursive Feature Elimination method to select the most important and effective features.

In a series of experiments, we trained a variety of machine learning algorithms. The results demonstrate a 76% accuracy and 83% precision in predicting cybersickness in the subjects based on the head movements. This study's contribution to the cybersickness literature lies in offering a preliminary analysis of a new source of data and providing insight into the relationship of head movements and cybersickness.

## 1    Introduction

In recent years, virtual reality (VR) has gained popularity as a technology that can provide users with immersive, interactive experiences. Various fields, such as entertainment [1], education[2], training [3], and healthcare [4], have found applications in virtual reality due to its ability toi simulate realistic environments and provide a sense of presence. Despite the benefits of VR, it has some limitations including the experience of cybersickness, a form of motion sickness, that may occur in virtual environments [5,

*These authors contributed equally to this work.

6]. Cybersickness prevents people from widely embracing and enjoying VR experiences due to symptoms such as nausea, dizziness, disorientation, and fatigue [7]. Additionally, research has shown differences between VR and face-to-face (F2F) conditions across various parameters, highlighting the importance of understanding these distinctions in optimizing user experiences [8, 9, 10, 11].

Cybersickness can be caused by a variety of factors, including those from hardware [12], software [13,14], users' individual differences, and the user's task [5, 15, 16]. While the exact mechanism is not yet fully modelled, it includes factors such cue conflict (when the sensory cues received by the brain from the virtual environment contradict the cues from the body's own senses) [17] and postural instability (how well a person can maintain balance) [1, 18]. This present research focuses on a specific form of postural sway which is head movements. Research shows that individuals who experience dizziness and patients with a vestibular deficiency show different patterns of head movements [19, 33]. Therefore, in this research we aim to use the head movements signals as a potential identifier for cybersickness. Head movements data can be a reliable, cheap, and easy to access source of data coming from accelerometer and gyroscope sensors embedded in all commercial VR headsets.

Several researchers have explored methods of predicting cybersickness using machine learning [19], which has shown promising results. By identifying patterns and appropriate identifiers extracted from physiological data; machine learning models can be trained to predict cybersickness [20, 21]. However, to the best of our knowledge, there has not been a study that utilizes and processes head movements data to train a machine learning model to make predictions for cybersickness.

This study develops a predictive model that provides predictive capability for cybersickness symptoms. This approach uses head movements data and its derivatives: velocity, acceleration, and jerk. Extensive feature extraction techniques are then applied to extract a wide range of statistical features, temporal features (time domain), and spectral features (frequency domain). To mitigate the problem of having a relatively small data sample, a feature selection technique called Recursive Feature Elimination [22] was used to choose only the most useful features in each experiment. A variety of machine learning models were run based on these features to test against a hold-out test set.

The next section of this paper provides an overview of related work on machine learning to identify cybersickness. The experimental methodology for the preprocessing, and extraction of features from this Polhemus dataset is described in [23]. A Polhemus is a 3D tracking device worn by the participants of the study that captured the dataset. The machine learning algorithms used for prediction are described, as well as their strengths and weaknesses. The paper then assesses the performance of the predictive model and discusses the implications of our findings, potential applications, and future research directions.

## 2 Related Work

Many researchers have explored methods of measuring cybersickness. While SSQ [24] has frequently been used as a subjective measure, self-reported by participants, researchers have widely explored the possibility of objective physiological measures, including EEG [25] ; EDA [26] ; heart rate, heart rate variability, blood pressure [27], [28], and respiration [27]. Unfortunately, results have been mixed. Since sensors for these methods can be expensive and feel invasive, popular headsets are not equipped with them. On the other hand, movement accelerometers are already included in most VR headsets. Previous research in the field has attempted to use postural or head movement data collected by the headset to identify cybersickness [20], but results have been inconsistent.

Various machine learning methods have been employed to predict cybersickness, and these approaches incorporated different kind of variables into their models. Certain studies have focused on internal physiological measurements, including factors such as heart rate [9], and muscle activity [28]. In addition, some investigations emphasize eye-related variables, including measures such as eye blinks and tracking eye movements [29, 32, 33]. Also, a subset of research concentrates on neural activities, employing electroencephalogram (EEG) data to drive their models [28]. Studies utilizing EEG data have reported promising results [34].

Among factors influencing cybersickness, body posture could be a significant contributor [1]. A sensory mismatch between what the eyes see and what the body feels in terms of balance and movement can disrupt the sensory integration process, causing the symptoms of cybersickness [35, 36]. Several studies have focused on this connection indicating that individuals with greater postural instability are more prone to experiencing severe cybersickness [1]. For example, adopting a standing posture, characterized by decreased stability, has been associated with a higher likelihood of severe sickness compared to a sitting posture. Also, regarding the locomotion methods, when the body movement is not well aligned with the movement in the VR environment, it can lead to cybersickness.

Early investigations by researchers like [37] and [1] highlighted a potential link, suggesting that both increases and decreases in body sway could be indicators of how individuals react to VR environments. These studies put forward the idea that monitoring our body's adjustments in VR could be key to predicting occurrences of cybersickness, highlighting the significance of postural changes in understanding our interactions with virtual spaces.

Contrasting viewpoints from studies such as those by [20] and [21] introduced a layer of skepticism, questioning the straightforward relationship between postural sway and cybersickness. These pieces of research pointed out instances where posture changes did not consistently match up with cybersickness episodes, raising doubts about the reliability of postural data as a catch-all predictor for VR-induced discomfort.

The debate extends further when focusing on specific movements, like those of the neck and back, with some researchers proposing a strong connection to cybersickness. However, findings from [38] countered this by showing that detailed analyses on these body parts failed to conclusively link them to cybersickness, illustrating the complexity of establishing a direct correlation. This diversity in research outcomes underscores the challenge in achieving a unified understanding of how our physical responses are intertwined with our virtual experiences.

The discussion around postural sway and cybersickness has focused mainly on using tools like force plates or analyzing overall body movements to measure postural stability. These approaches have provided valuable insights but also reveal a gap in our understanding of how specific parts of the body, particularly the head, contribute to the experience of cybersickness in virtual reality (VR) environments. Until now, the specific impact of head movements on cybersickness has not been extensively explored in the realm of cybersickness research.

This research introduces a novel perspective by specifically examining head movement data to understand its role in predicting cybersickness. This approach diverges from previous studies that have not fully considered the potential of head movements as a predictive measure for VR-induced discomfort. By shifting the focus to the head, we aim to uncover new insights into how the orientation and motion of the head could be closely linked with the onset of cybersickness. This novel approach not only fills a significant gap in the current research landscape but also opens up new avenues for developing more effective predictive models of cybersickness, potentially leading to improved VR design and user experiences.

## 3      Method

Figure 1 provides an overall view of the methods and approach in this study. Section 3.1 describes the dataset used in this research and discusses some details about the collection process and the experiment. Section 3.2 describes different forms of kinematics data that are calculated from the raw movement data and discusses the logic for incorporating these forms of signals. Section 3.3 explains the approach in windowing the data and discuss the benefits provided by it for our modeling performance. Section 3.4 discusses the feature extraction approach and explain the logic for choosing the wide range of features. And finally, Section 3.5 describes the modelling pipeline.

Innovative Cybersickness Detection: Exploring Head Movement Patterns in Virtual Reality

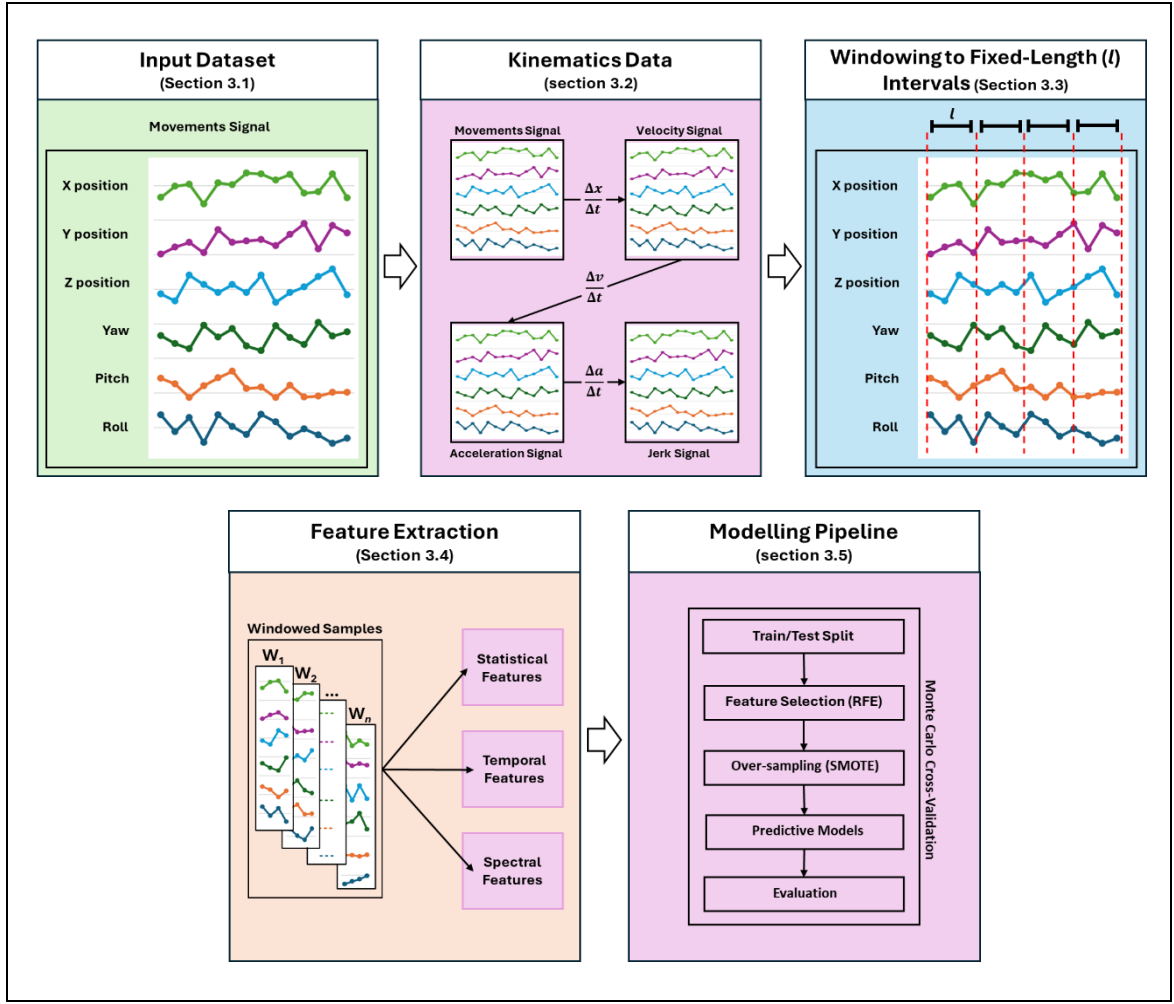

**Fig. 1.** Overall view of the study with representative diagrams of data; these are not real data.

### 3.1 Dataset

This study uses a secondary, publicly available dataset titled "APAL 2019: Postural Data, Game Performance, and Subjective Responses of Cybersickness in Virtual Reality Head-Mounted Displays" published by [16]. It encompasses data from 79 participants (41 women and 38 men) aged between 18 to 49, with an average height of 1.72 meters and weight of 71.58 kilograms. The study utilized the Simulator Sickness Questionnaire (SSQ) for assessing motion sickness, categorizing participants into "Well" or "Sick" groups based on their responses. The recorded variables include positional data (X, Y, Z in centimeters) and attitude data (pitch, roll, yaw in degrees) for the head. Due to the variability of SSQ data and its subjective nature, a different variable was chosen as dependent variable in our study. The last 10s of the movement signals of the participants who either indicated they were sick at the end of the experiment or chose to stop the experiment early because of severe sickness served as the "Sick" labeled samples. The first 10 seconds of all participants served as "not Sick" samples. Using this approach, the modeling task become a binary classification task. For more details about the dataset, please take a look at [16].

### 3.2    Kinematics Data

The recorded head movements signal in this study records six variables. It includes head disposition in centimeter along the X, Y, and Z axes and head rotations in degrees along three axes; pitch (flexion/extension), roll (lateral flexion), and yaw (axial rotation) (see Figure 2).

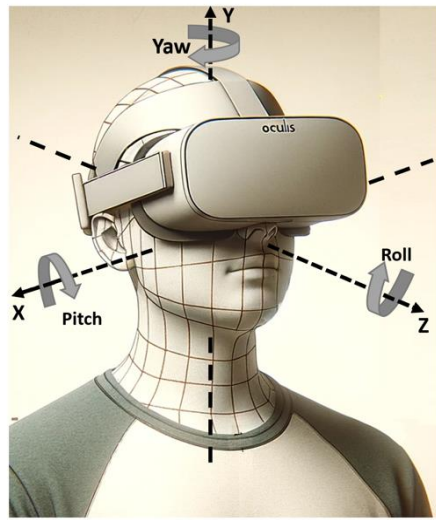

**Fig. 2.** Axes of the head movements in our study

In addition to the movements data, in the study of biomechanics and postural sway, researchers also consider the velocity and acceleration of the movements signal when doing kinematic analysis.[39, 40]

Jerk has also been used in several research as an indicator of postural sway smoothness, and postural instability [41, 42]. Jerk describes how smoothly or abruptly a movement changes in terms of acceleration. Higher values of jerk indicate more abrupt, and lower values of jerk indicate smoother movements.

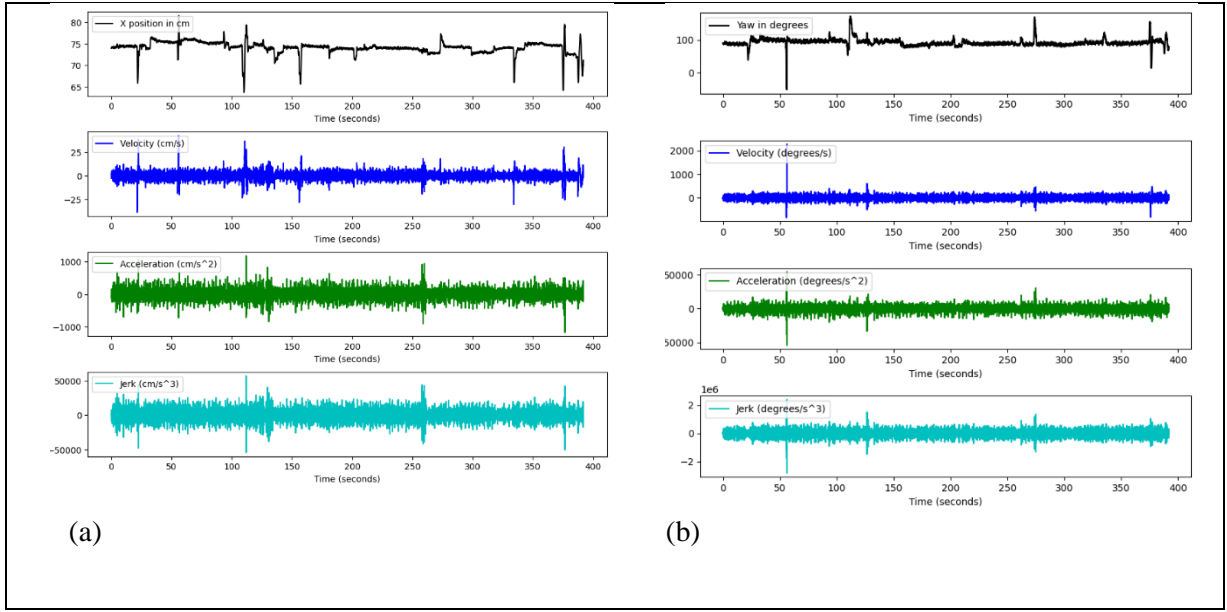

**Figure 3.** Derivatives of the original signals. (a) X position and its derivatives: Velocity (cm/s), Acceleration (cm/s^2), and Jerk (cm/s^3). (b) Yaw in degrees and its derivatives: Velocity (cm/s), Acceleration (cm/s^2), and Jerk (cm/s^3).

### 3.3 Windowing the Signals

As the first step to start processing the data, we perform a signal windowing task using a sliding window. Sliding window is a popular technique when dealing with time-series data. In this technique, the signals are divided into fixed length intervals, chosen by the user. This technique provided several benefits in the context of our research.

First, it provides practicality. Participants have participated in a 15-minute study. If one had chosen to use of the whole data to train the sickness detection model, the model could have made predictions only after at least after 15 minutes of observation, limiting the practicality and near-real time use cases of the model. Windowing the data into smaller lengths (1s, 2s, 3s, 5s, etc.) resulted in a model that made estimations of the sickness more frequently.

Second, windowing helps reduce the dimensionality and increasing a smaller sample size. When using machine learning algorithms, a large number of features and small number of samples can lead the model into issues like overfitting and poor generalization. In this present context, the original signal was a 15-minute time series of six variables recorded with 60hz sampling, resulting a dimensionality of 324,000 (15 mins x 60s x 60Hz x 6 variables) for only 75 observations. When considering a window size like 1s, the dimensionality would be reduced to 360 and number of observations would increase to 67,500.

Third, windowing improves the quality of the feature extraction methods. In this context, participants' movement signals, including temporal trend trajectories and spectral characteristics, may change over time, which is associated with the transition of the postural sway from a healthy state to the sickness state. By segmenting the signal into smaller, temporally localized intervals, windowing not only facilitates a more granular analysis of time-varying trends, but also significantly enhances frequency resolution.

### 3.4    Feature Extraction

Handling time series data in machine learning can be challenging as there are not any explicit feature coming with the raw data. As a result, a meticulous feature extraction process is needed to identify a set of appropriate features. In the literature of postural sway and Center of Position analysis using force plate data, researchers have put forward a vast set of variables that can be extracted from the data [43], [44], [45], [46]. These features can be mainly categorized in to three categories: statistical, temporal (time domain), and spectral (frequency domain).

While these lists are extensive and have been shown to be very effective in measuring a subjects' postural stability and balance, they are not directly applicable to the context of head movement data. To the best of our knowledge, there is only a limited body of literature discussing the processing and feature extraction of head movements data. These papers usually focus on a set of simple statistical features such as average acceleration, average velocity, and frequency of movements [39], [47]. As a result of the lack of literature in head movements signal processing, in this paper, a brute force approach was used. Influenced by postural sway literature, a large number of features in the three statistical, temporal, and spectral domains were used for the four types of kinematic data mentioned in Section 3.2.

It should be noted that while these features are statistically and mathematically meaningful, not all of them are good predictors of cybersickness. In addition, having a large number of these features can reduce the performance of our machine learning model. Therefore, we employ a feature selection technique, we select a handful of most meaningful features to train our model.

#### Statistical Features

When analyzing head movement data for detecting cybersickness, it was crucial to extract features that capture the complexity and nuances of such movements across different domains. Statistical features like kurtosis, skewness, mean, standard deviation, and histogram are fundamental as they provide insights into the distribution and variability of head movement data, enabling the identification of outliers or patterns indicative of cybersickness. These features help in understanding the basic structure and dispersion of the data, which are essential in distinguishing normal head movements from those affected by cybersickness.

#### Temporal Features

Time domain features, including autocorrelation, zero crossing rate, and mean absolute difference, were vital for capturing the temporal characteristics of head movements. These features allow for the analysis of how head movement changes over time, identifying specific patterns or irregularities that occur during episodes of cybersickness. For instance, autocorrelation helps in understanding the periodicity of movements, while neighborhood peaks can indicate sudden changes in movement direction or speed, which are common in cybersickness episodes.

**Spectral Features**

Time series data may include spectral and frequency information. A common method to look into spectral domain of the data employed by researchers in the study of Center of Position (COP) and postural sway is Fast Fourier Transform (FFT) [48, 49] . We note that applying FFT in the context of postural sway should be done with caution, as the postural sway and COP signals usually demonstrate nonstationary characteristics [49, 50]. However, windowing the data into smaller intervals can make the data closer to the stationary assumption.

In addition to Fourier transform techniques, the literature on postural sway analysis also recommends employing wavelet analysis for its distinct advantages. Wavelet transforms offer a more fitting approach for analyzing non-stationary signals due to their capability to capture intermittent, time-localized dynamics. This attribute makes them particularly effective for examining nonlinear systems that exhibit time delays [50].

Table 1 shows a list of all of the extracted features. We used a time series analysis Python package TSFEL to compute these features for each window [51].

**Table 1.** List of extracted features

| Feature Category | List of Features |
| --- | --- |
| Statistical Features | Absolute energy, Average power, ECDF, ECDF Percentile, ECDF Percentile Count, Entropy, Histogram, Interquartile range, Kurtosis, Max, Mean, Mean absolute deviation, Median, Median absolute deviation, Min, Peak to peak distance, Root mean square, Skewness, Standard deviation, Variance |
| Temporal Features | Area under the curve, Autocorrelation, Centroid, Mean absolute diff, Mean diff, Median absolute diff, Median diff, Negative turning points, Neighborhood peaks, Positive turning points, Signal distance, Slope, Sum absolute diff, Zero crossing rate |
| Spectral Features | FFT mean coefficient, Fundamental frequency, Wavelet absolute mean, Wavelet energy, Wavelet entropy, Wavelet standard deviation, Wavelet variance, Human range energy, LPCC, MFCC, Max power spectrum, Maximum frequency, Median frequency, Power bandwidth, Spectral centroid, Spectral decrease, Spectral distance, Spectral entropy, Spectral kurtosis, Spectral positive turning points, Spectral roll-off, Spectral roll-on, Spectral skewness, Spectral slope, Spectral spread, Spectral variation |

### 3.5    Modeling

The modeling process began by dividing the dataset into training and testing sets. To prevent data leakage from the training set to the test set, the participants were segregated into training and test groups before shuffling all windowed data. This approach guaranteed that no participant's data appears in both training and test sets, thus avoiding the potential for the model to learn participant-specific movement patterns. A stratified random sampling method [52] for the train/test split to maintain the proportionality of different classes in both sets.

In the second step, we used Recursive Feature Elimination [53] to identify the top 50 most informative and relevant features from our original dataset, which may contain up to 984 features per kinematic signal (i.e., 984, 1968, 2,952, or 3,936 features in each of the four experimental setting). RFE systematically assesses the importance of each feature, removing those with minimal contribution to the model's predictive accuracy. Given the small sample size in this study, RFE was particularly beneficial as it reduces model complexity and improves the performance [22, 53] . We trained a random forest-based RFE model on the training dataset and applied the same feature selection process to the test set.

In the third phase, the class imbalance issue was addressed within the training set using the Synthetic Minority Over-sampling Technique (SMOTE)[54] This method mitigates model bias towards the majority class by artificially augmenting the minority class.

Finally, to ensure the robustness and reliability of our results, a Monte Carlo Cross-Validation method [55] was used, repeating each experiment 50 times. This approach helps mitigate potential biases arising from random train/test splits, ensuring the stability and generalizability of our findings.

## 4    Results

Six machine learning models were trained on the processed training data and evaluated against the test set. These models include Logistic Regression, Random Forest, SVM, K-nearest Neighbors, Decision Tree, and an ensemble model Gradient Boosting. the modeling pipeline was tested in four settings, adding kinematics signals one by one. Table 2 shows the average results of the 50 repetitions of Monte Carlo cross validation experiments. Numbers in parentheses reflect the standard deviations. In all settings, Gradient Boosting outperformed other models in terms of accuracy, precision, and F1 Score.

Comparing different settings' results uncovered an interesting pattern. The performance of the best predictive model increased when more variations of the original movement signal were provided to the pipeline. Since all of the models in all the settings used the same number of 50 features selected by RFE, it could be assumed that the RFE choice of the most useful features should be different in different settings. Table 3 shows evaluation of this assumption. Also, it was observed that the standard deviation of the results decreased when more kinematic signals were added into our base feature set.

**Table 2.** Mean Results of Monte Carlo CV (50 repetition)

| Experiment | Models | Accuracy (SD) | Precision (SD) | Recall (SD) | F1 Score (SD) |
|---|---|---|---|---|---|
| Movement | Logistic Regression | 54.3% (8.6%) | 53.1% (11.3%) | 48.5% (19.2%) | 49.8% (14.7%) |
| | Random Forest | 58.0% (9.6%) | 64.0% (19.4%) | 30.5% (18.3%) | 39.7% (19.2%) |
| | **Gradient Boosting** | **63.4% (9.3%)** | **69.5% (13.2%)** | 44.6% (18.6%) | **58.0% (16.7%)** |
| | SVM | 58.8% (8.0%) | 60.3% (9.6%) | 47.7% (17.2%) | 52.2% (13.3%) |
| | K-Nearest Neighbors | 59.6% (7.9%) | 60.1% (9.0%) | **54.0% (16.7%)** | 56.0% (12.6%) |
| | Decision Tree | 61.5% (7.1%) | 64.9% (11.0%) | 47.0% (14.8%) | 53.8% (13.1%) |
| Movement + Velocity | Logistic Regression | 53.3% (9.3%) | 51.4% (14.2%) | 47.4% (20.1%) | 48.4% (16.7%) |
| | Random Forest | 64.7% (8.8%) | 76.6% (14.2%) | 41.6% (14.8%) | 52.9% (14.9%) |
| | **Gradient Boosting** | **72.7% (7.8%)** | **80.8% (9.9%)** | **60.1% (13.9%)** | **68.0% (10.9%)** |
| | SVM | 59.5% (8.0%) | 61.8% (12.6%) | 45.8% (15.3%) | 51.9% (13.7%) |
| | K-Nearest Neighbors | 56.3% (6.3%) | 55.3% (5.7%) | 64.8% (13.6%) | 59.2% (8.3%) |
| | Decision Tree | 62.1% (6.8%) | 65.1% (11.9%) | 48.6% (13.2%) | 55.2% (12.7%) |
| Movement + Velocity + Acceleration | Logistic Regression | 54.3% (8.0%) | 53.7% (10.5%) | 44.0% (18.2%) | 47.4% (14.6%) |
| | Random Forest | 66.1% (7.9%) | 77.7% (11.0%) | 44.3% (14.6%) | 55.4% (13.8%) |
| | **Gradient Boosting** | **74.5% (6.6%)** | **81.5% (7.0%)** | 63.6% (12.6%) | **70.8% (9.4%)** |
| | SVM | 60.5% (7.7%) | 65.5% (10.6%) | 42.1% (16.5%) | 50.0% (14.2%) |
| | K-Nearest Neighbors | 56.9% (6.1%) | 55.4% (5.1%) | **67.5% (14.0%)** | 60.5% (8.3%) |
| | Decision Tree | 63.2% (6.5%) | 66.7% (7.2%) | 51.3% (13.1%) | 57.4% (10.5%) |
| Movement + Velocity + Acceleration + Jerk | Logistic Regression | 55.6% (9.0%) | 55.0% (12.3%) | 46.6% (18.0%) | 49.6% (15.3%) |
| | Random Forest | 68.2% (8.1%) | 80.2% (10.5%) | 47.9% (14.6%) | 59.1% (13.5%) |
| | **Gradient Boosting** | **76.0% (5.9%)** | **82.7% (6.9%)** | 66.3% (11.5%) | **73.0% (8.1%)** |
| | SVM | 62.0% (7.7%) | 66.5% (10.8%) | 46.3% (16.2%) | 53.5% (14.1%) |
| | K-Nearest Neighbors | 58.3% (5.6%) | 56.3% (5.1%) | **72.0% (13.0%)** | 62.8% (8.1%) |
| | Decision Tree | 64.3% (6.7%) | 67.4% (7.3%) | 54.0% (12.9%) | 59.5% (10.7%) |

To examine the validity of our assumption that "RFE choice of the most useful features should be different in different settings" we look into the importance of the features of gradient boosting model which was the best model. Looking at the Table 3, we can see that indeed the RFE model is picking different set of features in different experiments when presented with additional, possibly more informative set of features.

**Table 3.** Results

| Experiment | Features Importance |
|---|---|
| Movement | ('movement_Roll _Neighborhood peaks', 0.101), ('movement_Roll _Fundamental frequency', 0.065), ('movement_Roll _Entropy', 0.062), ('movement_Roll _ECDF Percentile_1', 0.032), ('movement_X _Max', 0.03), ('movement_Z _ECDF Percentile_0', 0.0241), ('movement_Yaw _Neighbourhood peaks', 0.0199), ('movement_Z _Median', 0.016), ('movement_X _ECDF Percentile_1', 0.015), ('movement_Z _FFT mean coefficient_0', 0.013), ('movement_Roll _Mean', 0.013), ('movement_Roll _Max', 0.011), ('movement_Pitch _Spectral distance', 0.011), ('movement_Pitch _Wavelet standard deviation_6', 0.011), ('movement_Z _Mean', 0.01), ('movement_Y _Spectral distance', 0.01022Mean absolute diff', 0.0014), ('movement_Yaw _Autocorrelation', 0.0013), ('movement_Roll _Wavelet variance_3', 0.0004), ('movement_Y _Spectral entropy', 0.00041) |
| Movement + Velocity | ('Velocity_Z _Histogram_6', 0.0783), ('movement_Roll _Fundamental frequency', 0.0737), ('movement_Roll _Neighbourhood peaks', 0.0439), ('Velocity_Pitch _Histogram_6', 0.0255), ('movement_Roll _ECDF Percentile_1', 0.0228), ('movement_X _Max', 0.0201), ('movement_Z _ECDF Percentile_1', 0.017994609145126163), ('movement_Roll _Entropy', 0.0179), ('movement_Z _ECDF Percentile_0', 0.0170), ('movement_Pitch _Wavelet variance_6', 0.0146), ('movement_Z _Median', 0.0145), ('Velocity_Z _Histogram_8', 0.0110), ('Velocity_Z _Histogram_4', 0.0107), ('movement_Z _Absolute energy', 0.0104), ('movement_Z _FFT mean coefficient_0', 0.0102), ('Velocity_Roll _Histogram_9', 0.0102), ('movement_Pitch _Wavelet standard deviation_6', 0.0098), ('Velocity_Pitch _Histogram_4', 0.0097), ('Velocity_Pitch _Spectral roll-on', 0.0091), ('movement_X _ECDF Percentile_1', 0.0089), ('Velocity_Roll _Histogram_0', 0.0085) |
| Movement + Velocity + Acceleration | 'movement_Roll _Fundamental frequency', 0.0726), ('Velocity_Z _Histogram_6', 0.057), ('movement_Roll _Entropy', 0.0440), ('movement_Roll _Neighbourhood peaks', 0.037), ('acceleration_X _Neighbourhood peaks', 0.0313), ('acceleration_Z _Entropy', 0.0309), ('Velocity_Pitch _Spectral roll-on', 0.0273), ('movement_Roll _ECDF Percentile_1', 0.021), ('movement_Z _ECDF Percentile_1', 0.0157), ('movement_X _Max', 0.0125), ('movement_Yaw _Fundamental frequency', 0.0112), ('movement_Pitch _Wavelet standard deviation_6', 0.0108), ('acceleration_Pitch _Power bandwidth', 0.0105), ('movement_Roll _Max power spectrum', 0.0101), ('movement_Z _Median', 0.0098), ('movement_X _ECDF Percentile_1', 0.0097), ('movement_Z _ECDF Percentile_0', 0.0096), ('Velocity_Z _Histogram_4', 0.0094), ('movement_Pitch _Wavelet variance_6', 0.0087), |
| | ('Velocity_Yaw _Spectral roll-on', 0.0086), ('movement_Y _Power bandwidth', 0.0085), ('Velocity_Pitch _Histogram_1', 0.0084) |
| Movement + Velocity + Acceleratio + Jerk | ('movement_Roll _Fundamental frequency', 0.0870), ('Velocity_Z _Histogram_6', 0.0518), ('movement_Roll _Neighbourhood peaks', 0.0423), ('acceleration_Z _Entropy', 0.0276), ('Velocity_Pitch _Spectral roll-on', 0.0275), ('movement_Roll _Entropy', 0.0213), ('jerk__Y _Neighbourhood peaks', 0.0181), ('acceleration_X _Neighbourhood peaks', 0.01690), ('jerk__X _Neighbourhood peaks', 0.0157), ('movement_Roll _ECDF Percentile_1', 0.0151), ('movement_Yaw _Fundamental frequency', 0.0148), ('movement_Z _ECDF Percentile_1', 0.0141), ('movement_X _Max', 0.0130), ('movement_Z _Average power', 0.0128), ('movement_Pitch _Wavelet standard deviation_5', 0.0107), ('movement_Z _Median', 0.0106), ('acceleration_Z _Neighbourhood peaks', 0.0103), ('movement_Y _Power bandwidth', 0.0102), ('movement_Roll _Max power spectrum', 0.0097) |

# 5    Discussion

This study has introduced a novel approach to addressing the persistent issue of cybersickness in Virtual Reality (VR) environments. While cybersickness has been extensively studied through traditional physiological markers, this research focused on the analysis of head movement patterns as a promising marker for cybersickness detection. The study leveraged a publicly available dataset from a VR experiment involving 75 participants [23] and meticulously analyzed head movements and their derivatives across multiple axes and extracted statistical, temporal, and spectral features. Results included 76% accuracy and 83% precision in predicting cybersickness, showcasing the potential effectiveness of head movement patterns as an essential contributor to the understanding and mitigation of cybersickness in VR applications.

This study contributes to the field of cybersickness by providing an analysis of a new source of physiological marker that can be used as an indicator of cybersickness. The results also identify most valuable set of head movement features that are worth being extracted. The authors hope that future work can include further exploration of head movement data, combined with eye movement data to create real-time cybersickness detection systems.